\documentclass[11pt]{article}
\usepackage{acl}
\usepackage{times}
\usepackage{latexsym}
\usepackage[T1]{fontenc}
\usepackage[utf8]{inputenc}
\usepackage{microtype}
\usepackage{inconsolata}
\usepackage{graphicx}
\usepackage{amsmath}
\usepackage{amssymb}
\usepackage{booktabs}
\usepackage{multirow}
\usepackage{array}
\usepackage{xcolor}
\usepackage{placeins}
\usepackage{flafter}

\newcolumntype{L}[1]{>{\raggedright\arraybackslash}p{#1}}
\newcolumntype{C}[1]{>{\centering\arraybackslash}p{#1}}

\renewcommand{\topfraction}{0.85}
\renewcommand{\dbltopfraction}{0.85}
\renewcommand{\textfraction}{0.12}
\renewcommand{\floatpagefraction}{0.75}
\renewcommand{\dblfloatpagefraction}{0.75}
\newcommand{\dnaihao}[1]{}
\newcommand{\Yiming}[1]{}
\newcommand{\cyl}[1]{}
\newcommand{\prm}{LLaDA-PRM}

\title{LLaDA-PRM: A Bidirectional Step-Level Reasoning Evaluator}
\author{
  Yiming Feng$^{1}$ \quad Naihao Deng$^{1}$ \quad Yulong Chen$^{2,3}$ \quad Rada Mihalcea$^{1}$ \\
  $^{1}$University of Michigan \\
  $^{2}$University of Aberdeen \quad $^{3}$University of Cambridge\\
  \texttt{\{eliotfen,dnaihao,mihalcea\}@umich.edu} \\
  \texttt{s02yc6@abdn.ac.uk}
}
\begin{document}
\maketitle
\begin{abstract}
Step-level reasoning evaluators are commonly based on autoregressive language models, whose causal attention restricts each step representation to the problem, previous steps, and the current step. Yet, when the complete solution is available, the validity of an earlier step may become clearer only through its downstream consequences. We validate this hypothesis through a controlled 54-run comparison of causal and bidirectional LLaDA evaluators at 1B--3B scale, changing only the self-attention mask, and find bidirectional attention yields consistent improvements. Building on this finding, we introduce \prm{}, an 8B bidirectional evaluator that reaches 88.8 step-level F1 on MR-MATH-invalid and 83.8 on the out-of-distribution MR-GSM8K original-question subset, outperforming ReasonEval-Llemma-34B by 11.3 and 10.3 F1 points, respectively. \prm{} also remains effective when evaluating incomplete reasoning traces in online settings, outperforming the strongest baselines on both benchmarks by a large margin. We further show that \prm{} provides an effective training-data selection signal, improving Mistral-7B performance on MATH-500.

\end{abstract}

\section{Introduction}\label{sec:intro}

\begin{figure*}[t]
\centering
\includegraphics[width=0.92\textwidth]{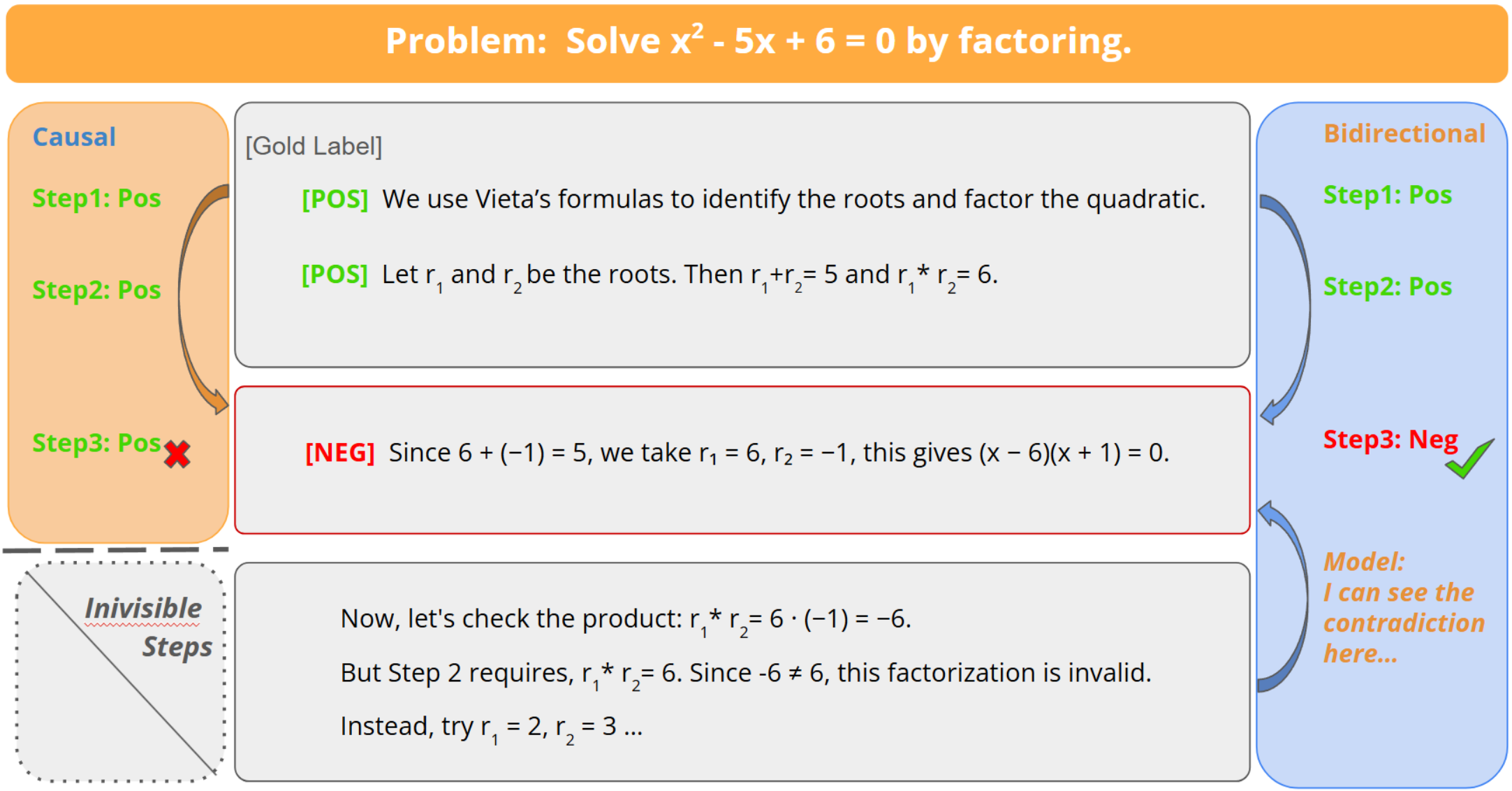}
\caption{Causal and bidirectional evaluators have access to different contextual information on the same step.}
\label{fig:later-step-schematic}
\end{figure*}

Large language models have made substantial progress on mathematical reasoning, especially when they generate intermediate reasoning steps before producing a final answer \citep{wei2022chain,lewkowycz2022solving,cobbe2021gsm8k,hendrycks2021math}.
As these reasoning traces become longer and more complex, it becomes increasingly important to evaluate whether each intermediate step is valid, useful, or erroneous, beyond final answer accuracy \citep{uesato2022solving,lightman2024lets,xia2024reasoneval,golovneva2022roscoe}.
Step-level reasoning evaluators, often instantiated as process reward models (PRMs), provide this finer-grained supervision and have become central to process supervision, reasoning-trace analysis, solution reranking, and inference-time search \citep{lightman2024lets,wang2024mathshepherd,chen2024autoprm,luo2024omegaprm,zhang2025generative}.

Most existing step-level evaluators are built by fine-tuning autoregressive language models, including Mistral-, WizardMath-, and Llemma-based evaluators used in ReasonEval-style and PRM-style systems \citep{xia2024reasoneval,wang2024mathshepherd,jiang2023mistral,luo2023wizardmath,azerbayev2023llemma}.
However, autoregressive models adopt causal attention, and thus the representation of a step is computed from the problem, the preceding steps, and the current step, but not the later reasoning steps. 
In a completed-solution evaluation, later steps may provide evidence about earlier ones. For instance, a step that appears locally plausible may later lead to an inconsistency, failed derivation, or self-correction. As shown in Figure~\ref{fig:later-step-schematic}, an incorrect factorization (``$(x{-}6)(x{+}1) = 0$'') initially appears plausible with the preceding reasoning steps, because it satisfies the sum-of-roots constraint. However, later steps like (``$6 \cdot (-1) = -6 \neq 6$'') expose a contradiction through the product-of-roots calculation. 
By contrast, humans naturally use later evidence when judging the earlier step \citep{resulaj2009changes, van2016common}.

Motivated by such human patterns, we first compare leveraging later steps (bidirectional attention) versus not leveraging later steps (causal attention) in a controlled setup.
We train matched causal and bidirectional step-level evaluators from scratch on PRM800K~\citep{lightman2024lets} while keeping the model architecture, data, and optimization recipe fixed.
Across various model sizes, bidirectional evaluators consistently outperform causal evaluators in 22 of 27 matched comparisons, with consistent gains in step-level accuracy, macro-F1, and macro-AUC. Thus, bidirectional attention is more effective than causal attention for full-trace step-level evaluation under matched training conditions.

We also scale the same design with LLaDA-8B-Base \citep{nie2025llada}, a pretrained diffusion language model with built-in bidirectional attention.
The resulting \prm{} achieves step-level F1 scores of 88.8 on MR-MATH-invalid~\citep{xia2024reasoneval} and 83.8 in the out-of-distribution MR-GSM8K original-question subset~\citep{zeng2024mrgsm8k}, outperforming the strongest published ReasonEval baseline on each benchmark by 11.3 and 10.3 F1 points, respectively~\citep{xia2024reasoneval}.

Under online evaluation, where each step is scored using only the problem and the steps generated so far, \prm{} remains effective and outperforms the strongest published baselines on MR-MATH-invalid and the MR-GSM8K original-question subset.
Furthermore, a Mistral-7B model trained on a $5{,}222$-example red-only subset selected by \prm{} outperforms a model trained on a size-matched random subset by $+3.4$ percentage points on MATH-500.
Overall, \prm{} establishes state-of-the-art results on the two error-detection benchmarks and provides a useful signal for downstream training-data selection.

Our main contributions are threefold: we isolate the effect of bidirectional attention through a controlled comparison with causal attention, introduce \prm{}, and demonstrate its effectiveness in online evaluation and downstream data filtering.

\section{Related Work}\label{sec:related}

\paragraph{Step-level reasoning evaluators and process reward models.}
Process supervision evaluates intermediate reasoning steps rather than only the final answer, providing finer-grained feedback for mathematical reasoning \citep{uesato2022solving,lightman2024lets}.
PRMs assign scores to individual steps and have been used for solution selection, search, and training-time supervision~\citep{lightman2024lets,wang2024mathshepherd,luo2024omegaprm}. 
Recent work studies how to collect such process-level signals and use them for training and inference, including Math-Shepherd, AutoPRM, OmegaPRM, and generative verifiers \citep{wang2024mathshepherd,chen2024autoprm,luo2024omegaprm,zhang2025generative}.
A complementary line of work evaluates completed reasoning traces and detects step-level errors. ReasonEval annotates completed solutions with positive, neutral, and negative step labels \citep{xia2024reasoneval}, while ROSCOE, ProcessBench, and PRMBench further study reasoning-trace evaluation and step-level error detection \citep{golovneva2022roscoe,zheng2025processbench,song2025prmbench}. 
In contrast, our \prm{} uses built-in bidirectional attention and achieves state-of-the-art results on the two error-detection benchmarks considered here.

\paragraph{Using later evidence for step assessment.}
Several recent methods recognize that information beyond the current and preceding steps can be useful for judging reasoning quality. OVM and BiRM incorporate future-success or value-style signals while still relying on causal backbones \citep{yu2024ovm,chen2025better}. BiPRM scores reasoning traces in both left-to-right and right-to-left prompt orders and combines the resulting scores at the output level \citep{zhang2025biprm}. These approaches share the intuition that later reasoning can inform the assessment of earlier steps. However, they do not change the self-attention pattern inside the backbone. Each forward pass still represents a step under a directional attention constraint. In contrast, we evaluate the completed trace with bidirectional self-attention in a single forward pass, allowing each scored step to condition on later reasoning steps.
\paragraph{Bidirectional backbones for reasoning evaluation.}
Bidirectional attention has long been standard for classification and representation learning in encoder models \citep{devlin2019bert, liu2020roberta, he2021debertav3}.
LLaDA~\citep{nie2025llada}, which uses a masked diffusion language-modeling objective and built-in bidirectional attention, scales to 8B parameters and matches autoregressive models on language generation.
Because LLaDA is pretrained to recover masked tokens using surrounding context on both sides, its representations are naturally suited to settings where the full input sequence is available.

\section{Method}\label{sec:method}
We first formulate the step-level evaluation task and then describe the step representations and training objective used by our evaluators.
\subsection{Task Formulation}
\label{sec:task}
Given a math problem $q$ and a completed solution split into ordered steps $\hat{h}_{1:n}$, we formulate a three-way probability vector for each supervised step:
\[
\mathbf{p}_i =
(p_i^{\mathrm{neg}}, p_i^{\mathrm{neu}}, p_i^{\mathrm{pos}}).
\]
The \textit{pos} label means that the step is correct and useful, \textit{neu} means that the step is valid but redundant, and \textit{neg} means that the step contains a calculation or logical error. 
This three-way formulation allows us to define separate validity and redundancy scores for a more fine-grained assessment.
Following \citet{xia2024reasoneval}, we define:
\begin{equation}
\label{eq:scores}
\begin{aligned}
s_i^{\mathrm{val}} &= p_i^{\mathrm{pos}} + p_i^{\mathrm{neu}}, \\
s_i^{\mathrm{red}}   &= p_i^{\mathrm{neu}}, \\
S_{solution}^{\mathrm{val}}   &= \min_i s_i^{\mathrm{val}}, \\
S_{solution}^{\mathrm{red}}     &= \max_i s_i^{\mathrm{red}} .
\end{aligned}
\end{equation}
For the benchmark evaluation, we follow \citet{xia2024reasoneval} and use fixed binary decision thresholds at both the step and solution levels. At the step level, a step is judged valid if $s_i^{\mathrm{val}} \geq 0.5$ and redundant if $s_i^{\mathrm{red}} \ge 0.15$. At the solution level, we apply the same thresholds to the aggregated scores: a solution is judged valid if $S_{solution}^{\mathrm{val}} \geq 0.5$ and redundant if $S_{solution}^{\mathrm{red}} \ge 0.15$.

\subsection{Step Representations}
\label{sec:mask-representations}

Let $x_{1:T}$ denote the tokenized input sequence formed from the problem $q$ and the completed reasoning trace $\hat{h}_{1:n}$.
Let $e_i$ be the token position corresponding to the end of step $\hat{h}_i$.
We compare two variants of the same backbone that differ only in attention.
In the causal variant, each position can attend only to itself and earlier positions.
In contrast, for the bidirectional variant, each position can attend to the full input sequence.
In additive-mask notation, where position $u$ attends to key position $v$, the two masks are\begin{equation}
\begin{aligned}
M^{\mathrm{full}}_{uv} &= 0 \qquad \forall u,v,\\
M^{\mathrm{causal}}_{uv} &=
\begin{cases}
0, & v \le u,\\
-\infty, & v > u,
\end{cases}
\end{aligned}
\end{equation}
The corresponding step representations are
\begin{equation}
\begin{aligned}
\mathbf{h}^{\mathrm{bidir}}_i
&= \left[f_\theta(x_{1:T};M^{\mathrm{full}})\right]_{e_i}
   \equiv g_\theta^{\mathrm{bidir}}(x_{1:T}),\\
\mathbf{h}^{\mathrm{causal}}_i
&= \left[f_\theta(x_{1:T};M^{\mathrm{causal}})\right]_{e_i}
   \equiv g_\theta^{\mathrm{causal}}(x_{1:e_i}).
\end{aligned}
\end{equation}
$\mathbf{h}^{\mathrm{causal}}_i$ is computed from the problem, the preceding steps, and the current step $\hat{h}_i$, whereas $\mathbf{h}^{\mathrm{bidir}}_i$ can also encode later reasoning steps $\hat{h}_{i+1:n}$.

\subsection{Step-Level Scoring and Training}
\label{sec:bidir-step-eval}

We replace the backbone's language-modeling head with a linear score head $W \in \mathbb{R}^{3 \times d}$ that maps each step representation to a three-class logit vector, followed by a softmax:
\begin{equation}
\mathbf{z}_i = W \mathbf{h}_i, \qquad
\mathbf{p}_i = \mathrm{softmax}(\mathbf{z}_i).
\end{equation}
The resulting $\mathbf{p}_i \in \mathbb{R}^3$ is the three-way probability vector from \S\ref{sec:task}.

Let $\mathcal{S}$ be the set of supervised step-end positions in a training example, with gold labels $y_i \in \{\textit{neg}, \textit{neu}, \textit{pos}\}$. We train the model with cross-entropy loss averaged over these positions:
\begin{equation}
\mathcal{L} = -\frac{1}{|\mathcal{S}|} \sum_{i \in \mathcal{S}} \log p_i^{y_i}.
\end{equation}

\begin{table*}[!t]
\centering\small
\setlength{\tabcolsep}{4pt}
\renewcommand{\arraystretch}{1.08}
\begin{tabular}{@{}l ccc ccc ccc@{}}
\toprule
                & \multicolumn{3}{c}{\textbf{1B}} & \multicolumn{3}{c}{\textbf{2B}} & \multicolumn{3}{c}{\textbf{3B}} \\
\cmidrule(lr){2-4}\cmidrule(lr){5-7}\cmidrule(lr){8-10}
\textbf{Architecture} & Acc & F1 & AUC & Acc & F1 & AUC & Acc & F1 & AUC \\
\midrule
LLaDA-bidir   & \textbf{45.4} & 40.1 & 59.6 & \textbf{46.9} & \textbf{41.1} & \textbf{60.8} & \textbf{46.7} & \textbf{41.1} & \textbf{60.6} \\
LLaDA-causal  & 45.1 & \textbf{40.4} & \textbf{59.7} & 45.2 & 39.9 & 59.6 & 44.1 & 38.7 & 58.5 \\
\midrule
$\boldsymbol{\Delta}$ LLaDA-bidir $-$ LLaDA-causal (pp) 
              & $+0.3$ & $-0.3$ & $-0.1$ & $\boldsymbol{+1.7}$ & $\boldsymbol{+1.2}$ & $\boldsymbol{+1.2}$ & $\boldsymbol{+2.6}$ & $\boldsymbol{+2.4}$ & $\boldsymbol{+2.1}$ \\
\bottomrule
\end{tabular}

\caption{Step-level test metrics on the 3{,}000-trace subset, averaged over 3 seeds. Metric values are reported as percentages, and deltas are reported in percentage points. The full 54-run grid is reported in Appendix~\ref{sec:appendix_sweep3}.}

\label{tab:main-3k-results}
\end{table*}

\section{Bidirectional versus Causal Attention}
\label{sec:ablation}
PRMs traditionally use causal autoregressive backbones, restricting each step representation to preceding context. We therefore compare causal and bidirectional attention under matched training conditions.

\subsection{Setup}\label{sec:ablation-setup}

\paragraph{Models.}
We use the LLaDA Transformer architecture~\citep{nie2025llada} and construct \textit{from-scratch} models at 1B, 2B, and 3B parameters, to eliminate potential influence from model pre-training.
For each size, the bidirectional variant uses LLaDA's native bidirectional attention, while the causal variant adopts the causal attention that can only attend to the current and earlier positions (\S\ref{sec:mask-representations}).
At each size, both variants use the same tokenizer, architectural dimensions, initialization scheme, and step-level scoring head.
The only difference between the two variants is the attention mask that creates causal and bidirectional attention.

\paragraph{Data.}
We train and evaluate on PRM800K~\citep{lightman2024lets}, a step-annotated collection of GPT-4-generated solutions to MATH problems~\citep{hendrycks2021math}.
Because the natural step-label distribution is heavily skewed toward positive labels in both train and test splits, we construct separate label-balanced diagnostic training and test splits for the controlled ablation; details are given in Appendix~\ref{sec:appendix_repro}.
We sample 3 train sets of 1{,}000, 2{,}000, and 3{,}000 traces from the balanced train split and a test set of 608 traces from the test split.
The same subsets are used for the bidirectional and causal variants.

\paragraph{Training and evaluation.}
For every combination of the two attention masks, three model sizes, and three training-set sizes, we train three replicas with different random seeds, for a total of $54$ runs.
Within each matched cell, the bidirectional and causal runs share the same data subset, random seed, and optimization hyper-parameters.
We report step-level accuracy, macro-F1, and macro-AUC in percentages.
A matched comparison pairs the bidirectional and causal runs with the same model size, data size, and seed, giving $27$ seed-level pairs for comparison.
Full architecture and training details for the controlled grid are provided in Appendix~\ref{sec:appendix_sweep3_train}.

\subsection{Results}
\label{sec:ablation-mask}
\paragraph{At 1B, the two attention patterns are effectively tied.}
Table~\ref{tab:main-3k-results} reports step-level metrics on the 3{,}000-trace subset.
At 1B scale, LLaDA-bidir reaches $45.4$ accuracy, $40.1$ macro-F1, and $59.6$ macro-AUC, while LLaDA-causal reaches $45.1$ accuracy, $40.4$ macro-F1, and $59.7$ macro-AUC.
The corresponding bidirectional-minus-causal deltas are therefore $+0.3$ pp accuracy, $-0.3$ pp macro-F1, and $-0.1$ pp macro-AUC.
These differences are all within $0.3$ pp, so the 1B setting does not show a clear advantage for either attention mask.

\paragraph{At 2B, bidirectional attention improves all three metrics.}
At 2B scale, LLaDA-bidir reaches $46.9$ accuracy, $41.1$ macro-F1, and $60.8$ macro-AUC.
The matched LLaDA-causal model reaches $45.2$ accuracy, $39.9$ macro-F1, and $59.6$ macro-AUC.
Thus, the bidirectional variant improves over the causal variant by $+1.7$ pp accuracy, $+1.2$ pp macro-F1, and $+1.2$ pp macro-AUC.
Unlike the 1B case, all three cells favor bidirectional attention, indicating that the full-trace advantage becomes visible at this scale.

\paragraph{At 3B, the bidirectional advantage becomes larger.}
At 3B scale, LLaDA-bidir reaches $46.7$ accuracy, $41.1$ macro-F1, and $60.6$ macro-AUC.
LLaDA-causal reaches $44.1$ accuracy, $38.7$ macro-F1, and $58.5$ macro-AUC.
The resulting gains are $+2.6$ pp accuracy, $+2.4$ pp macro-F1, and $+2.1$ pp macro-AUC, which are larger than the corresponding 2B gains on every metric.
The 3B results therefore show that, on the largest from-scratch model in this controlled grid, bidirectional attention gives a consistent and stronger improvement over causal attention.

\paragraph{The full matched grid shows a broad bidirectional advantage.}
Across the full matched grid, LLaDA-bidir outperforms LLaDA-causal in $22$ of $27$ paired cells for accuracy, $22$ of $27$ paired cells for macro-F1, and $22$ of $27$ paired cells for macro-AUC. The mean bidirectional-minus-causal differences are $+2.03$ pp accuracy, $+0.90$ pp macro-F1, and $+1.05$ pp macro-AUC. Appendix~\ref{sec:appendix_sweep3} reports the full grid and analysis. Taken together, these controlled results show that, under matched training conditions, bidirectional attention is more effective than causal attention for full-trace step-level evaluation.

\section{\prm{}: A Pretrained Bidirectional Evaluator}\label{sec:scale}
Building on the controlled comparison above, we scale the bidirectional evaluator to LLaDA-8B-Base and introduce \prm{}. We evaluate it on full-trace benchmarks and under online evaluation without later-step evidence.

\subsection{Setup}\label{sec:scale-setup}
\paragraph{Modeling.}
We build \prm{} by fine-tuning LLaDA-8B-Base~\citep{nie2025llada}. LLaDA uses bidirectional self-attention through masked diffusion language modeling, so the backbone can condition on the full input sequence in a single forward pass. This matches our full-trace evaluation setting, where the evaluator scores each intermediate step in the context of a completed reasoning trace. Following \S\ref{sec:bidir-step-eval}, we keep LLaDA's bidirectional self-attention, replace its language-modeling head with a three-class linear score head, and train with cross-entropy loss at the supervised step-end positions.

\paragraph{Datasets.}
We fine-tune \prm{} on PRM800K~\citep{lightman2024lets}, the step-annotated dataset introduced in \S\ref{sec:ablation-setup}.
Full training-set details are provided in Appendix~\ref{sec:appendix_8b_train}.

For evaluation we use three meta-evaluation benchmarks. MR-MATH-invalid, released by~\citet{xia2024reasoneval}, contains $159$ MATH solutions, all of which reach the correct final answer and $83$ of which contain an annotated mid-trace reasoning error. It tests whether the evaluator can identify reasoning errors when the final-answer outcome looks correct. MR-GSM8K is built the same way on GSM8K~\citep{cobbe2021gsm8k} solutions and is therefore out-of-distribution with respect to our MATH-based training data. It tests whether the evaluator generalizes across math problem distributions. The main comparison uses its $1{,}418$-example original-question subset, matching the published ReasonEval baselines; Appendix~\ref{sec:appendix_gsm8k_full} reports the full $2{,}999$-example set. MR-MATH-redundant~\citep{xia2024reasoneval} contains MATH solutions whose intermediate steps include valid but unnecessary computation. It tests whether the evaluator distinguishes redundant steps from incorrect ones.

\paragraph{Training Details.}
We fine-tune \prm{} for one epoch with an effective batch size of $8$ and a maximum sequence length of $2{,}048$ tokens. The training runs on $4\times$ RTX~6000 Pro GPUs with FSDP. Appendix~\ref{sec:appendix_training} shows more details.

\paragraph{Baselines.} 
WizardMath-V1.1-7B~\citep{luo2023wizardmath} and Llemma-34B~\citep{azerbayev2023llemma} are math-oriented base models that ReasonEval~\citep{xia2024reasoneval} fine-tuned on PRM800K. Among the published ReasonEval baselines included here, ReasonEval-WizardMath-V1.1 and ReasonEval-Llemma-34B provide the strongest results at the 7B and 34B scales. We also include Math-Shepherd-Mistral-7B~\citep{wang2024mathshepherd}, an open-source process reward model, via the score file released by \citet{xia2024reasoneval}.

We further added two controlled comparisons under our own pipeline.
First, we fine-tune WizardMath-V1.1-7B with our PRM800K data recipe, effective batch size, scoring head, and evaluation code, giving a causal-backbone control under matched pipeline implementation conditions.
Second, we fine-tune a batch-matched LLaDA-bidir control with effective batch size $64$, matching the batch size used by the published ReasonEval baselines while otherwise following our \prm{} data recipe and evaluation pipeline. The corresponding control configurations are summarized in Appendix~\ref{sec:appendix_8b_controls}.

\begin{table*}[t]
\centering\footnotesize
\setlength{\tabcolsep}{4pt}
\renewcommand{\arraystretch}{1.08}
\begin{tabular}{@{}L{0.35\textwidth}C{0.07\textwidth}C{0.10\textwidth}C{0.10\textwidth}C{0.10\textwidth}C{0.10\textwidth}@{}}
\toprule
\textbf{Model} & \textbf{Params} & \textbf{Sol-F1} & \textbf{Sol-AUC} & \textbf{Step-F1} & \textbf{Step-AUC} \\
\midrule
\multicolumn{6}{@{}l}{\textbf{MR-MATH-invalid (in-domain)}} \\
Math-Shepherd & 7B & 70.1 & 77.3 & 60.0 & 77.2 \\
ReasonEval-Mistral & 7B & 78.0 & 85.1 & 68.6 & 85.7 \\
ReasonEval-WizardMath-V1.1 & 7B & 78.6 & 87.5 & 73.9 & 89.5 \\
WizardMath-V1.1  (control, eff\_bs=8)  & 7B & 72.5 & 79.6 & 64.7 & 82.1 \\
ReasonEval-Llemma & 34B & 79.6 & 90.8 & 77.5 & 92.8 \\
Batch-matched LLaDA-bidir (eff\_bs=64) & 8B & 85.9 & 94.8 & 79.3 & 92.8 \\
LLaDA-PRM (eff\_bs=8, headline) & 8B & \textbf{93.0} & \textbf{97.2} & \textbf{88.8} & \textbf{96.4} \\
\midrule
\multicolumn{6}{@{}l}{\textbf{MR-GSM8K (out-of-distribution)}} \\
ReasonEval-Mistral & 7B & 61.8 & 79.8 & 62.9 & 86.1 \\
ReasonEval-WizardMath-V1.1 & 7B & 74.1 & 90.7 & 72.8 & 91.4 \\
WizardMath-V1.1  (control, eff\_bs=8) & 7B & 68.0 & 81.5 & 66.2 & 82.2 \\
ReasonEval-Llemma & 34B & 81.0 & 88.1 & 73.5 & 86.8 \\
Batch-matched LLaDA-bidir (eff\_bs=64) & 8B & 76.4 & 88.3 & 72.6 & 87.8 \\
LLaDA-PRM (eff\_bs=8, headline) & 8B & \textbf{90.2} & \textbf{95.7} & \textbf{83.8} & \textbf{93.0} \\
\midrule
\multicolumn{6}{@{}l}{\textbf{MR-MATH-redundant (in-domain)}} \\
Math-Shepherd & 7B & 50.4 & 54.5 & 42.7 & 53.0 \\
ReasonEval-Mistral & 7B & 60.7 & 63.4 & 59.7 & 70.9 \\
ReasonEval-WizardMath-V1.1  & 7B & 61.6 & 64.8 & 59.7 & \textbf{72.2} \\
WizardMath-V1.1  (control, eff\_bs=8)  & 7B & 60.0 & 64.9 & 60.5 & \textbf{72.4} \\
ReasonEval-Llemma & 34B & 58.3 & 62.7 & 57.5 & 67.3 \\
Batch-matched LLaDA-bidir (eff\_bs=64) & 8B & 63.3 & 67.7 & 60.5 & 68.8 \\
LLaDA-PRM (eff\_bs=8, headline) & 8B & \textbf{63.6} & \textbf{68.4} & \textbf{61.0} & 71.2 \\
\bottomrule
\end{tabular}
\caption{Performance on the three benchmarks.}
\label{tab:benchmarks}
\end{table*}

\subsection{Results}\label{sec:scale-meta}

\paragraph{\prm{} exceeds the strongest baseline on MR-MATH-invalid.}
Table~\ref{tab:benchmarks} reports the full metric values on the three meta-evaluation benchmarks.
On MR-MATH-invalid, \prm{} reaches $93.0$ Sol-F1, $97.2$ Sol-AUC, $88.8$ Step-F1, and $96.4$ Step-AUC.
Compared with ReasonEval-Llemma, the strongest published baseline on this benchmark, these scores correspond to gains of $+13.4$ pp Sol-F1, $+6.4$ pp Sol-AUC, $+11.3$ pp Step-F1, and $+3.6$ pp Step-AUC.
These deltas show that the scaled bidirectional evaluator is especially strong on the main in-domain error-detection benchmark.

\paragraph{\prm{} generalizes better on the out-of-distribution benchmark.}
In the MR-GSM8K original-question subset, \prm{} reaches $90.2$ Sol-F1, $95.7$ Sol-AUC, $83.8$ Step-F1, and $93.0$ Step-AUC.
These scores improve over the strongest published baseline for each metric by $+9.2$ pp Sol-F1 and $+10.3$ pp Step-F1 over ReasonEval-Llemma, and by $+5.0$ pp Sol-AUC and $+1.6$ pp Step-AUC over ReasonEval-WizardMath-V1.1.
The largest margins again appear on solution- and step-level F1, suggesting that the scaled evaluator transfers well from the MATH-based training distribution to GSM8K-style reasoning traces.
The AUC gains are smaller, but \prm{} still outperforms the strongest published baselines on both AUC metrics, indicating that the improvement is not limited to a single thresholded metric. The main comparison uses the original-question subset; results on all $2{,}999$ MR-GSM8K examples, including the reversed and program-of-thought (\emph{POT}) variants, are reported in Appendix~\ref{sec:appendix_gsm8k_full}.

\paragraph{\prm{} remains competitive on MR-MATH-redundant.}
On MR-MATH-redundant, \prm{} reaches $63.6$ Sol-F1, $68.4$ Sol-AUC, $61.0$ Step-F1, and $71.2$ Step-AUC.
Relative to ReasonEval-WizardMath-V1.1, the strongest published baseline on this benchmark, \prm{} improves by $+2.0$ pp Sol-F1, $+3.6$ pp Sol-AUC, and $+1.3$ pp Step-F1, but trails by $1.0$ pp on Step-AUC.
This benchmark therefore shows a more mixed pattern: \prm{} is strongest on three of four headline metrics, while redundancy-sensitive Step-AUC remains a limitation.

\paragraph{Controlled comparisons show the same trend under matched conditions.}
Under our matched pipeline, \prm{} outperforms the WizardMath-V1.1 causal control by $+20.5$ pp Sol-F1 and $+22.2$ pp Sol-F1 on MR-MATH-invalid and MR-GSM8K respectively. On MR-MATH-redundant, \prm{} is higher on Sol-F1, Sol-AUC, and Step-F1, but the WizardMath-V1.1 control has the best Step-AUC.
The batch-matched LLaDA-bidir control also supports the in-domain trend: at effective batch size $64$, it reaches $85.9$ Sol-F1 on MR-MATH-invalid, remaining $+7.3$ pp above ReasonEval-WizardMath-V1.1.
Together, the published-baseline and controlled comparisons show that scaling the bidirectional evaluator to the pretrained LLaDA-8B backbone yields strong performance on error detection, while redundancy detection remains the least settled part of the evaluation.

\begin{table*}[!t]
\centering\small
\setlength{\tabcolsep}{6pt}
\renewcommand{\arraystretch}{1.08}
\begin{tabular}{@{}l l c c c c@{}}
\toprule
\textbf{Benchmark} & \textbf{Configuration} & \textbf{Sol-F1} & \textbf{Sol-AUC} & \textbf{Step-F1} & \textbf{Step-AUC} \\
\midrule
\multirow{5}{*}{MR-MATH-invalid}    & ReasonEval-WizardMath-V1.1 (7B) & $78.6$ & $87.5$ & $73.9$ & $89.5$ \\
                                     & ReasonEval-Llemma (34B)         & $79.6$ & $90.8$ & $77.5$ & $92.8$ \\
                                     & LLaDA-PRM (online)     & $\mathbf{89.8}$ & $\mathbf{96.4}$ & $\mathbf{84.9}$ & $\mathbf{94.7}$ \\
\cmidrule(lr){2-6}
                                     & LLaDA-PRM (offline)    & $93.0$ & $97.2$ & $88.8$ & $96.4$ \\
                                     & $\Delta$ online $-$ offline (pp) & $-3.2$ & $-0.8$ & $-3.9$ & $-1.7$ \\
\midrule
\multirow{5}{*}{MR-GSM8K (OOD)}      & ReasonEval-WizardMath-V1.1 (7B) & $74.1$ & $90.7$ & $72.8$ & $91.4$ \\
                                     & ReasonEval-Llemma (34B)         & $81.0$ & $88.1$ & $73.5$ & $86.8$ \\
                                     & LLaDA-PRM (online)     & $\mathbf{89.6}$ & $\mathbf{95.3}$ & $\mathbf{83.7}$ & $\mathbf{92.8}$ \\
\cmidrule(lr){2-6}
                                     & LLaDA-PRM (offline)    & $90.2$ & $95.7$ & $83.8$ & $93.0$ \\
                                     & $\Delta$ online $-$ offline (pp) & $-0.6$ & $-0.4$ & $-0.1$ & $-0.2$ \\
\midrule
\multirow{5}{*}{MR-MATH-redundant}   & ReasonEval-WizardMath-V1.1 (7B) & $\mathbf{61.6}$ & $64.8$ & $59.7$ & $\mathbf{72.2}$ \\
                                     & ReasonEval-Llemma (34B)         & $58.3$ & $62.7$ & $57.5$ & $67.3$ \\
                                     & LLaDA-PRM (online)     & $60.1$ & $\mathbf{66.6}$ & $\mathbf{60.7}$ & $70.3$ \\
\cmidrule(lr){2-6}
                                     & LLaDA-PRM (offline)    & $63.6$ & $68.4$ & $61.0$ & $71.2$ \\
                                     & $\Delta$ online $-$ offline (pp) & $-3.5$ & $-1.8$ & $-0.3$ & $-0.9$ \\
\bottomrule
\end{tabular}
\caption{Offline and online evaluation of LLaDA-PRM.}
\label{tab:online-vs-offline}
\end{table*}

\subsection{Online Evaluation}\label{sec:scale-mechanism}

\paragraph{Online evaluation removes later-step evidence.}
In addition to the setting where the completed trace is available, process reward models are often deployed in step-level beam search~\citep{wang2024mathshepherd}. The key property is that only the steps generated so far are available at scoring time. To test whether \prm{} remains accurate in this setting, for each labeled step $\hat{h}_i$ in a benchmark trace, we construct a separate input containing the problem and the steps generated so far, $(q,\hat{h}_{1:i})$, run the evaluator on that input, and read the score at $\hat{h}_i$. Thus, when scoring the current step, the model has no access to later reasoning steps.

\paragraph{Online performance drops modestly across benchmarks.}
The result also remains strong on MR-MATH-invalid, MR-GSM8K, and MR-MATH-redundant (Table~\ref{tab:online-vs-offline}).
Comparing the online and offline rows in Table~\ref{tab:online-vs-offline}, Sol-F1 drops by at most $3.5$\,pp, Sol-AUC by less than $2$\,pp, Step-F1 by less than $4$\,pp, and Step-AUC by less than $2$\,pp.
On MR-MATH-invalid, the largest metric drop is $3.9$\,pp Step-F1, while Sol-F1 drops by $3.2$\,pp (from $93.0$ to $89.8$). On MR-GSM8K, Sol-F1 drops by only $0.6$\,pp (from $90.2$ to $89.6$), and Step-F1 is essentially unchanged.
Even under online evaluation, \prm{} continues to outperform ReasonEval-Llemma-34B on the in-domain MR-MATH-invalid benchmark (Sol-F1 $89.8$ versus $79.6$) and on the out-of-distribution MR-GSM8K benchmark (Sol-F1 $89.6$ versus $81.0$).
On MR-MATH-redundant, online \prm{} leads on Sol-AUC ($66.6$ versus $64.8$) and Step-F1 ($60.7$ versus $59.7$), but trails ReasonEval-WizardMath-V1.1 on Sol-F1 ($60.1$ versus $61.6$) and Step-AUC ($70.3$ versus $72.2$).

\section{Data Filtering for SFT}\label{sec:filter}

Beyond meta-evaluation, we test whether \prm{}'s step-level scores are useful as a data-selection signal for supervised fine-tuning (SFT) on mathematical reasoning. Following the data-filtering protocol of \citet{xia2024reasoneval}, we use the evaluator to filter candidate solution traces before training.

\subsection{Setup}\label{sec:filter-setup}

\paragraph{Dataset.}
Our candidate pool is a random sample of $10{,}000$ MATH problem--solution pairs from MMIQC~\citep{liu2024mmiqc}. Filtering operates at the solution level. We score each candidate solution with \prm{} and ReasonEval-WizardMath-V1.1~\citep{xia2024reasoneval}. The Val + Red filter uses the validity and redundancy thresholds defined in Section~\ref{sec:method} and keeps solutions judged both valid and non-redundant, retaining $5{,}222$ solutions ($52\%$) under \prm{} and $4{,}437$ ($44\%$) under ReasonEval-WizardMath-V1.1. We add two single-dimension ablations at the same matched size: val-only keeps the top-$K$ solutions ranked by $S_{solution}^{\mathrm{val}}$, and red-only keeps the bottom-$K$ solutions ranked by $S_{solution}^{\mathrm{red}}$ (the solutions judged least redundant). The random arm is a uniform subsample at the same size.

\paragraph{Training.}
We fine-tune \texttt{Mistral-7B-v0.1}~\citep{jiang2023mistral}, a 7B general-purpose base language model that is not specialized for mathematical reasoning. We train with FSDP across 4 GPUs. Every configuration is trained with three random seeds. The complete hyperparameters are given in Appendix~\ref{sec:appendix_mistral_sft}.

\subsection{Results}\label{sec:filter-results}

\paragraph{\prm{} red-only filtering gives the strongest downstream signal.}
Table~\ref{tab:filter_ablation} reports the data-filtering results.
Within the \prm{} block, red-only reaches $22.5\%$ Acc with an average generation length of $249$ tokens, giving both the highest accuracy and the shortest generations.
Compared with the size-matched random subset, red-only improves Acc by $+3.4$\,pp ($22.5$ vs.\ $19.1$) and reduces generation length by $22$ tokens.
It also outperforms Val + Red by $+1.6$\,pp and Val-only by $+3.3$\,pp, while producing substantially shorter generations than both filters.
This indicates that the redundancy score provides the strongest data-selection signal in this experiment.

\paragraph{Validity filtering is weaker than redundancy filtering.}
Val-only reaches $19.2\%$ Acc, only $+0.1$\,pp above the random subset of the same size.
Val + Red is stronger, reaching $20.9\%$ Acc, but remains $1.6$\,pp below red-only and has the longest average generation length ($280$ tokens).
Thus, combining validity and redundancy is stronger than validity-only selection in this experiment, but remains weaker than redundancy-only selection.

\paragraph{The ReasonEval scorer does not improve over random selection.}
Under ReasonEval-WizardMath-V1.1, all three filtered subsets are below the corresponding random subset in accuracy.
Val + Red reaches $19.2\%$ Acc, while Val-only and red-only each reach $19.3\%$ Acc, compared with $20.2\%$ for random.
The red-only filter still produces the shortest generations ($250$ tokens), but this reduction does not translate into higher accuracy.
Overall, \prm{}'s scores provide a more useful downstream data-selection signal for Mistral-7B SFT than the published ReasonEval-WizardMath-V1.1 scores.

\begin{table}[t]
\centering
\small
\setlength{\tabcolsep}{3.5pt}
\renewcommand{\arraystretch}{1.08}
\begin{tabular*}{\columnwidth}{@{\extracolsep{\fill}}llrrr@{}}
\toprule
\textbf{Scorer} & \textbf{Filter} & \textbf{N} & \textbf{Acc.} & \textbf{Tok.} \\
\midrule
\multirow{3}{*}{\shortstack[l]{\prm{}}}
  & Val + Red & 5,222 & 20.9 & 280 \\
  & Val-only  & 5,222 & 19.2 & 276 \\
  & Red-only  & 5,222 & \textbf{22.5} & \textbf{249} \\
\midrule
\multirow{3}{*}{\shortstack[l]{ReasonEval\\WizardMath-V1.1}}
  & Val + Red & 4,437 & 19.2 & 276 \\
  & Val-only  & 4,437 & 19.3 & 267 \\
  & Red-only  & 4,437 & 19.3 & 250 \\
\midrule
  & Random    & 5,222 & 19.1 & 271 \\
  & Random    & 4,437 & 20.2 & 272 \\
\bottomrule
\end{tabular*}
\caption{Data-filtering ablation on Mistral-7B-v0.1, averaged over seeds. $N$ is the number of training samples, Acc. is MATH-500 greedy Pass@1 reported as a percentage, and Tok. is mean generation length.}
\label{tab:filter_ablation}
\end{table}

\section{Conclusion}\label{sec:conclusion}

We study bidirectional attention for full-trace step-level reasoning evaluation through a controlled attention-mask ablation. Using the LLaDA Transformer architecture, we train matched causal and bidirectional evaluators from scratch at 1B--3B scale while holding the architecture family, data subsets, random seeds, optimizer, training recipe, and scoring head fixed. Across the 54-run grid, the bidirectional evaluator wins $22$ of $27$ matched comparisons on accuracy, macro-F1, and macro-AUC, with consistent gains on all three metrics. These results provide controlled evidence that, when completed reasoning traces are available, bidirectional attention is more effective than causal attention for step-level evaluation under matched training conditions.

Motivated by this result, we fine-tune LLaDA-8B-Base as \prm{} and evaluate it on ReasonEval-style benchmarks. \prm{} reaches $88.8$ Step-F1 on MR-MATH-invalid and $83.8$ Step-F1 on the out-of-distribution MR-GSM8K original-question subset. Under online evaluation, where each step is scored using only the problem and the steps generated so far, performance drops modestly and remains above the strongest published baselines on the two error-detection benchmarks. Beyond meta-evaluation, \prm{}'s scores also serve as a useful data-selection signal for SFT, with redundancy-based filtering improving Mistral-7B accuracy on MATH-500 by $+3.4$\,pp over a random subset of the same size. Together, these findings support bidirectional attention as an effective design choice for full-trace reasoning evaluation, while showing that the resulting evaluator remains useful when later reasoning steps are unavailable.

\section*{Limitations}\label{sec:limitations}
\textbf{Architecture coverage.} The controlled decomposition varies only the self-attention mask within the LLaDA architecture. We do not test other discrete-diffusion backbones such as MDLM or SEDD, encoder-family bidirectional models such as BERT-style architectures, or LLaMA-style autoregressive backbones with causal attention. The broader architectural question remains open beyond LLaDA.

\textbf{Data and task coverage.} PRM800K is MATH-only, and out-of-distribution generalization is tested only on GSM8K-style problems.

\textbf{Neutral-class supervision and redundancy coverage.}
PRM800K's step-label distribution is heavily skewed toward \textit{pos}: \textit{neu} accounts for only $7.8\%$ of the labeled training steps and $6.6\%$ of the test steps, whereas \textit{pos} accounts for $79.6\%$ and $82.0\%$, respectively. Because \textit{neu} denotes valid but redundant steps, this imbalance provides substantially less supervision for distinguishing useful reasoning from valid but unproductive reasoning. Appendix~\ref{sec:appendix_imbalance} provides a detailed analysis of this label distribution. Models trained on PRM800K are therefore constrained by the limited neutral and redundancy-oriented supervision available in the dataset. 

\textbf{Attribution of 8B gains.} The controlled 1B--3B ablation supports the conclusion that bidirectional attention helps under matched training conditions. The 8B benchmark results, however, also depend on the released LLaDA-8B-Base pretrained weights and the diffusion language-model pretraining recipe behind that checkpoint. We therefore do not isolate how much of the 8B SOTA performance comes from bidirectional attention, model scale, pretrained weights, or diffusion-LM pretraining. The 8B bidirectional evaluator, the batch-matched 8B control, and the 7B causal control are also single-seed; multi-seed 8B replication would further strengthen this estimate.

\section*{Ethics Statement}
This work uses no human subjects. PRM800K \citep{lightman2024lets} is publicly released under a permissive license. Compute: training the 8B model used approximately 24 GPU-hours on RTX 6000 Pro and A40 hardware (precise figures in Appendix~\ref{sec:appendix_training}).
Process reward models are used to guide LLM training and inference-time selection; the architectural improvement we describe could be applied to existing PRM training pipelines. We see no specific harm pathway distinct from the broader RLHF concerns documented in prior work.

\bibliography{custom}

\appendix
\setcounter{topnumber}{3}
\setcounter{dbltopnumber}{2}
\setcounter{totalnumber}{5}
\renewcommand{\topfraction}{0.92}
\renewcommand{\dbltopfraction}{0.95}
\renewcommand{\textfraction}{0.06}
\renewcommand{\floatpagefraction}{0.82}
\renewcommand{\dblfloatpagefraction}{0.82}
\section{Training Details}\label{sec:appendix_training}
This appendix details training infrastructure for both the controlled 54-run sweep (\S\ref{sec:ablation}) and the pretrained 8B model (\S\ref{sec:scale}).

\subsection{Architecture Adjustments}
\label{sec:appendix_arch}
We construct three from-scratch LLaDA configurations at 1B, 2B, and 3B parameters by downscaling the LLaDA-8B-Base reference along four axes: hidden size $d_\text{model}$, number of layers $n_\text{layers}$, number of attention heads $n_\text{heads}$, and MLP hidden size. Table~\ref{tab:scratch-configs} lists the resulting dimensions with the 8B reference row included for comparison. All remaining components are held fixed at every size. Each model uses RMSNorm with $\epsilon=10^{-5}$, SiLU activations, rotary positional embeddings with $\theta=500{,}000$, an attention layout with $n_\text{kv\_heads}=n_\text{heads}$ (no grouped-query or multi-query attention), no QKV bias, and untied input and output embeddings. The tokenizer and vocabulary are inherited from LLaDA-8B-Base at $126{,}464$ tokens. The architectural maximum sequence length is $4{,}096$; training uses $1{,}024$.

\begin{table}[t]
\centering\small
\setlength{\tabcolsep}{5pt}
\begin{tabular}{@{}l rrrr@{}}
\toprule
\textbf{Config} & $\boldsymbol{d_\text{model}}$ & $\boldsymbol{n_\text{layers}}$ & $\boldsymbol{n_\text{heads}}$ & \textbf{MLP} \\
\midrule
LLaDA-8B-Base (ref.) & 4096 & 32 & 32 & 12288 \\
\midrule
LLaDA-1B & 2048 & 16 & 16 & 5504 \\
LLaDA-2B & 2560 & 24 & 20 & 6912 \\
LLaDA-3B & 2816 & 26 & 22 & 7168 \\
\bottomrule
\end{tabular}
\caption{Scratch-config dimensions for the 1B--3B grid, obtained by downscaling LLaDA-8B-Base along the four reported axes. All other architectural components are held fixed across sizes.}
\label{tab:scratch-configs}
\end{table}

For step-level evaluation we replace LLaDA's language-modeling head $\text{ff\_out}$ with a $d_\text{model} \to 3$ linear score head that produces the three-class logit vector defined in \S\ref{sec:bidir-step-eval}. The original $\text{ff\_out}$ is frozen and does not contribute to the optimizer state. Within each size, the LLaDA-bidir and LLaDA-causal arms share architecture, initialization, and training schedule, so the self-attention mask is the only variable between matched arms. LLaDA-bidir uses LLaDA's native full bidirectional attention bias. LLaDA-causal adds an upper-triangular additive bias with $-\infty$ entries above the diagonal, so each token attends only to itself and preceding tokens. The same head replacement is applied to the pretrained 8B backbone in \S\ref{sec:appendix_8b_train}, while the headline 8B model retains LLaDA's native bidirectional attention.

\subsection{Controlled 1B--3B grid}
\label{sec:appendix_sweep3_train}
The 54 training runs span the three from-scratch configurations from \S\ref{sec:appendix_arch}, two attention masks, three training-set sizes ($1{,}000$, $2{,}000$, $3{,}000$ traces), and three random seeds (13, 42, 7). Training uses a distributed world size of $6$ with FSDP full sharding, automatic wrapping, and no FSDP activation checkpointing. Per-device batch sizes are $4$, $4$, and $2$ for the 1B, 2B, and 3B models, respectively, with gradient accumulation fixed at $1$, giving effective batches of $24$, $24$, and $12$. The maximum sequence length is $1{,}024$ tokens. Phase~2 traces have a median length of approximately $200$ tokens, so the $1{,}024$-token limit accommodates $99.3\%$ of the training examples without truncation. The optimizer is AdamW ($\beta_1{=}0.9$, $\beta_2{=}0.999$, weight decay $0.01$). The learning rate is $5\!\times\!10^{-5}$ with a cosine schedule and an 8-step warmup; we clip the gradient norm at $1.0$ and use bf16 mixed precision. Each run trains for 3 epochs on its assigned subset. Initialization uses the truncated-normal Mitchell scheme via LLaDA's \texttt{reset\_parameters()}. Training data is shuffled once per epoch with the run's seed and then length-sorted for memory-stable batching.

\begin{figure*}[t]
\centering
\includegraphics[width=\textwidth]{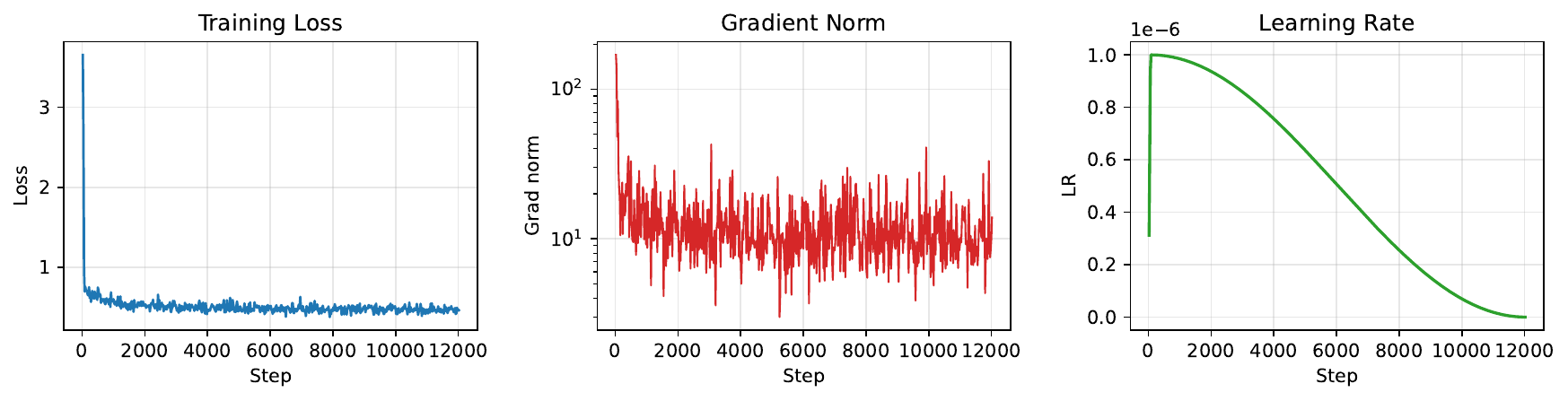}
\caption{\prm{} 8B fine-tuning over the single epoch on PRM800K Phase~2 ($\sim$12k optimizer steps). \textbf{Left:} training cross-entropy falls from $3.65$ to $\sim$0.5 and then plateaus. \textbf{Center:} gradient norm (log scale) stays bounded after the warmup spike. \textbf{Right:} the learning rate follows the $64$-step warmup and cosine decay. All curves are taken from the run's \texttt{log\_history.json}.}
\label{fig:8b-training}
\end{figure*}

\subsection{LLaDA-PRM 8B fine-tuning}
\label{sec:appendix_8b_train}
We fine-tune \prm{} from the released LLaDA-8B-Base checkpoint for one epoch on the full PRM800K Phase~2 training set with effective batch size $8$ and maximum sequence length $2{,}048$ tokens. The optimizer is AdamW ($\beta_1{=}0.9$, $\beta_2{=}0.95$, weight decay $0.1$) with a learning rate of $1\times 10^{-6}$; the learning rate follows a cosine schedule with $64$ warmup steps, and we clip the gradient norm at $0.5$. Training uses bf16 mixed precision and activation checkpointing, and the archived run uses four GPUs with FSDP. Its recorded runtime is $21{,}784.6$ seconds, corresponding to $24.2$ GPU-hours.

\paragraph{Training dynamics.}
The 8B fine-tune is stable and converges within its single epoch. Figure~\ref{fig:8b-training} shows the training cross-entropy dropping sharply over the first few hundred steps and then plateauing near $0.5$, with a gradient norm that stays bounded after the initial warmup spike and the intended warmup-then-cosine learning-rate schedule.

\subsection{Controls for the 8B benchmark}
\label{sec:appendix_8b_controls}
The 7B causal control fine-tunes the WizardMath-V1.1-7B checkpoint with the same core optimization settings as \prm{}: effective batch size $8$, maximum length $2{,}048$, AdamW settings, learning rate, 64-step warmup, bf16 precision, one epoch, and seed 13. It retains the model's native causal architecture and uses model-specific FSDP settings, a different save schedule, and $96{,}279$ rather than $96{,}280$ training examples. The LLaDA-bidir effective-batch-64 control fine-tunes the same LLaDA-8B-Base checkpoint using the paper-matched batch-64 recipe; it uses 8 warmup steps and different logging/save intervals. These runs support score-level control comparisons, but neither is a one-variable ablation of the headline run.

\subsection{Mistral-7B SFT for data filtering}
\label{sec:appendix_mistral_sft}
For the data-filtering experiments in \S\ref{sec:filter}, we fine-tune \texttt{Mistral-7B-v0.1} on each filtered MMIQC subset with FSDP across $4$ GPUs, using a 4-epoch cosine schedule and reporting the epoch-4 checkpoint. The effective batch size is $32$ (per-device $2$, gradient accumulation $4$) and the maximum sequence length is $2{,}048$. The optimizer is AdamW with a learning rate of $5\!\times\!10^{-7}$, a cosine schedule, and $16$ warmup steps. Every configuration is trained with three random seeds ($42$, $43$, $44$). Pass@1 on MATH-500 is graded with the math-verify grader.

\FloatBarrier

\section{Full 1B--3B Grid Results}\label{sec:appendix_sweep3}

This appendix reports all $54$ runs in the controlled 1B--3B sweep. Table~\ref{tab:full-grid-results} gives the per-run test metrics for every combination of attention mask (LLaDA-bidir or LLaDA-causal), model size (1B, 2B, or 3B), training-set size ($1{,}000$, $2{,}000$, or $3{,}000$ traces), and random seed (13, 42, or 7). Each row corresponds to one trained model; no seed averaging is applied in this table. Pairing each bidirectional run with the causal run that shares its model size, training subset, and seed yields $27$ seed-matched run pairs.

Table~\ref{tab:paired-stats} summarizes the $27$ seed-matched pairs descriptively. LLaDA-bidir wins $22$ of $27$ pairs for each step-level metric, with mean differences of $+2.03$ pp accuracy, $+0.90$ pp macro-F1, and $+1.05$ pp macro-AUC.

\begin{table*}[!t]
\centering\footnotesize
\setlength{\tabcolsep}{10pt}
\renewcommand{\arraystretch}{0.94}
\begin{tabular}{@{}ccccrrrr@{}}
\toprule
\textbf{Size} & \textbf{Data} & \textbf{Seed} & \textbf{Mask}
& \textbf{Acc} & \textbf{macro-F1} & \textbf{macro-AUC}
& \textbf{Late-AUC} \\
\midrule
1B & 1k & 13 & Bidir  & 48.1 & 41.4 & 61.8 & 31.8 \\
1B & 1k & 13 & Causal & 42.9 & 39.4 & 59.2 & 18.9 \\
1B & 1k & 42 & Bidir  & 45.7 & 39.9 & 60.9 & 33.3 \\
1B & 1k & 42 & Causal & 46.6 & 41.3 & 61.3 & 24.4 \\
1B & 1k & 7  & Bidir  & 45.9 & 41.1 & 61.8 & 35.0 \\
1B & 1k & 7  & Causal & 45.4 & 40.9 & 60.9 & 26.0 \\
1B & 2k & 13 & Bidir  & 45.7 & 39.1 & 59.1 & 17.6 \\
1B & 2k & 13 & Causal & 41.1 & 37.4 & 57.9 & 18.5 \\
1B & 2k & 42 & Bidir  & 44.5 & 40.4 & 61.4 & 25.8 \\
1B & 2k & 42 & Causal & 43.5 & 39.9 & 60.6 & 23.0 \\
1B & 2k & 7  & Bidir  & 43.0 & 38.8 & 59.3 & 19.4 \\
1B & 2k & 7  & Causal & 43.5 & 40.6 & 60.7 & 16.9 \\
1B & 3k & 13 & Bidir  & 44.9 & 39.9 & 59.3 & 13.7 \\
1B & 3k & 13 & Causal & 44.1 & 39.5 & 58.7 & 16.3 \\
1B & 3k & 42 & Bidir  & 44.8 & 39.1 & 59.2 & 16.5 \\
1B & 3k & 42 & Causal & 47.2 & 42.4 & 61.0 & 16.6 \\
1B & 3k & 7  & Bidir  & 46.4 & 41.3 & 60.4 & 18.9 \\
1B & 3k & 7  & Causal & 44.1 & 39.2 & 59.4 & 19.1 \\
\midrule
2B & 1k & 13 & Bidir  & 47.4 & 40.1 & 61.8 & 34.1 \\
2B & 1k & 13 & Causal & 43.5 & 39.7 & 59.9 & 22.4 \\
2B & 1k & 42 & Bidir  & 46.3 & 40.4 & 61.1 & 31.1 \\
2B & 1k & 42 & Causal & 42.2 & 38.5 & 59.0 & 18.7 \\
2B & 1k & 7  & Bidir  & 46.9 & 42.2 & 62.1 & 33.9 \\
2B & 1k & 7  & Causal & 44.9 & 40.6 & 60.5 & 21.3 \\
2B & 2k & 13 & Bidir  & 47.2 & 40.7 & 60.0 & 17.3 \\
2B & 2k & 13 & Causal & 45.4 & 40.9 & 60.3 & 18.2 \\
2B & 2k & 42 & Bidir  & 46.1 & 40.5 & 60.5 & 21.0 \\
2B & 2k & 42 & Causal & 42.2 & 39.0 & 59.4 & 20.4 \\
2B & 2k & 7  & Bidir  & 44.2 & 39.7 & 60.0 & 17.6 \\
2B & 2k & 7  & Causal & 41.9 & 39.1 & 59.4 & 15.4 \\
2B & 3k & 13 & Bidir  & 47.1 & 41.2 & 60.1 & 16.0 \\
2B & 3k & 13 & Causal & 46.5 & 40.0 & 59.4 & 12.2 \\
2B & 3k & 42 & Bidir  & 45.5 & 40.0 & 60.2 & 20.1 \\
2B & 3k & 42 & Causal & 44.1 & 39.7 & 59.4 & 18.9 \\
2B & 3k & 7  & Bidir  & 48.1 & 42.0 & 62.0 & 22.5 \\
2B & 3k & 7  & Causal & 44.9 & 40.1 & 60.0 & 13.8 \\
\midrule
3B & 1k & 13 & Bidir  & 50.6 & 43.8 & 62.6 & 29.4 \\
3B & 1k & 13 & Causal & 43.6 & 40.0 & 60.3 & 17.6 \\
3B & 1k & 42 & Bidir  & 45.4 & 39.6 & 60.3 & 27.2 \\
3B & 1k & 42 & Causal & 43.5 & 39.4 & 60.7 & 22.9 \\
3B & 1k & 7  & Bidir  & 44.3 & 40.9 & 62.2 & 32.9 \\
3B & 1k & 7  & Causal & 46.2 & 41.3 & 60.3 & 18.9 \\
3B & 2k & 13 & Bidir  & 46.1 & 40.3 & 60.3 & 24.8 \\
3B & 2k & 13 & Causal & 42.7 & 38.5 & 58.2 & 12.8 \\
3B & 2k & 42 & Bidir  & 47.3 & 41.4 & 60.1 & 17.7 \\
3B & 2k & 42 & Causal & 45.9 & 40.5 & 59.4 & 13.9 \\
3B & 2k & 7  & Bidir  & 43.7 & 39.6 & 59.8 & 20.8 \\
3B & 2k & 7  & Causal & 42.6 & 38.6 & 58.2 & 13.0 \\
3B & 3k & 13 & Bidir  & 46.9 & 40.9 & 60.2 & 19.2 \\
3B & 3k & 13 & Causal & 43.7 & 38.3 & 58.7 & 12.2 \\
3B & 3k & 42 & Bidir  & 48.3 & 41.7 & 60.7 & 23.2 \\
3B & 3k & 42 & Causal & 43.1 & 37.4 & 57.8 & 11.1 \\
3B & 3k & 7  & Bidir  & 44.9 & 40.8 & 61.0 & 20.6 \\
3B & 3k & 7  & Causal & 45.4 & 40.4 & 59.1 & 10.6 \\
\bottomrule
\end{tabular}
\caption{Per-run step-level test results for all $54$ runs in the controlled grid. Each row represents one trained model evaluated on the same held-out 608-trace test set. All metric values are percentages and are rounded to one decimal place for display; paired comparisons use the unrounded values. Bidir and Causal rows with the same model size, training-set size, and seed form one matched pair.}
\label{tab:full-grid-results}
\end{table*}

\FloatBarrier

\begin{figure*}[t]
\centering
\includegraphics[width=\textwidth]{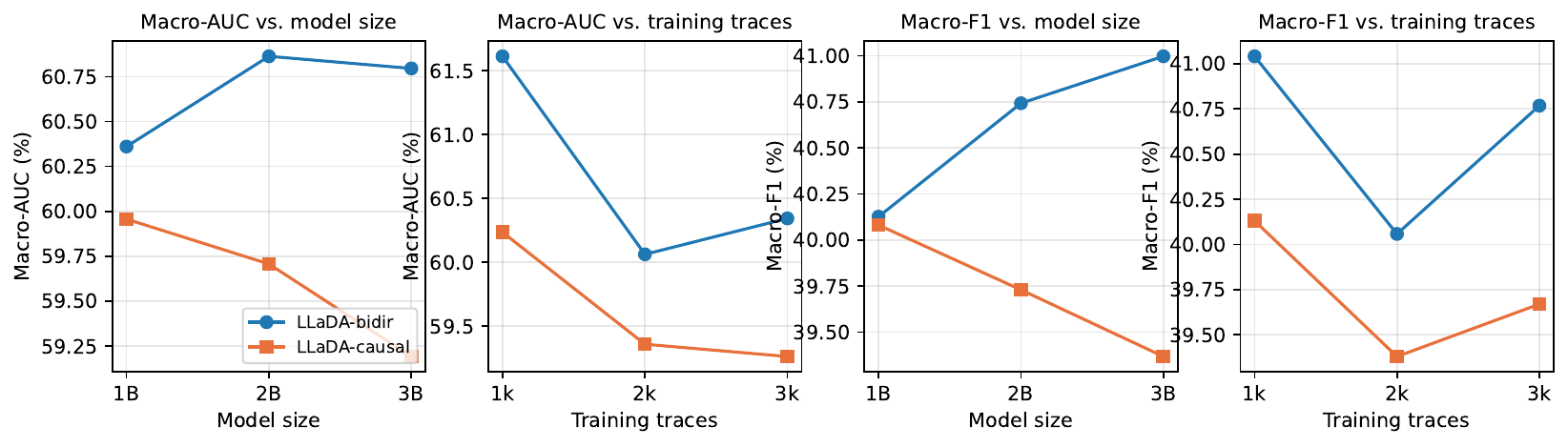}
\caption{Step-level macro-AUC and macro-F1 versus model size (1B/2B/3B) and training-set size (1k/2k/3k), averaged over the other factors and the three seeds. LLaDA-bidir (blue) is at or above LLaDA-causal (orange) in every panel.}
\label{fig:grid-scaling}
\end{figure*}

\begin{table}[!t]
\centering
\small
\setlength{\tabcolsep}{6pt}
\renewcommand{\arraystretch}{1.08}
\begin{tabular}{@{}lcc@{}}
\toprule
\textbf{Metric} & \textbf{Wins} & \textbf{Mean $\Delta$ (pp)} \\
\midrule
Accuracy  & $22/27$ & $+2.03$ \\
macro-F1  & $22/27$ & $+0.90$ \\
macro-AUC & $22/27$ & $+1.05$ \\
\bottomrule
\end{tabular}
\caption{Descriptive LLaDA-bidir $-$ LLaDA-causal contrasts over the $27$ seed-matched run pairs. Mean differences are computed from unrounded metrics.}
\label{tab:paired-stats}
\end{table}

\paragraph{The advantage becomes clearer at larger model sizes.}
Figure~\ref{fig:grid-scaling} tracks macro-AUC and macro-F1 as functions of model size and training-set size. The bidirectional arm is at or above the causal arm in every panel, and the macro-F1 gap widens with model size: the two arms are indistinguishable after rounding at 1B but separate clearly by 3B. This pattern is consistent with the bidirectional advantage becoming clearer at larger model sizes.

\begin{figure}[t]
\centering
\includegraphics[width=0.84\columnwidth]{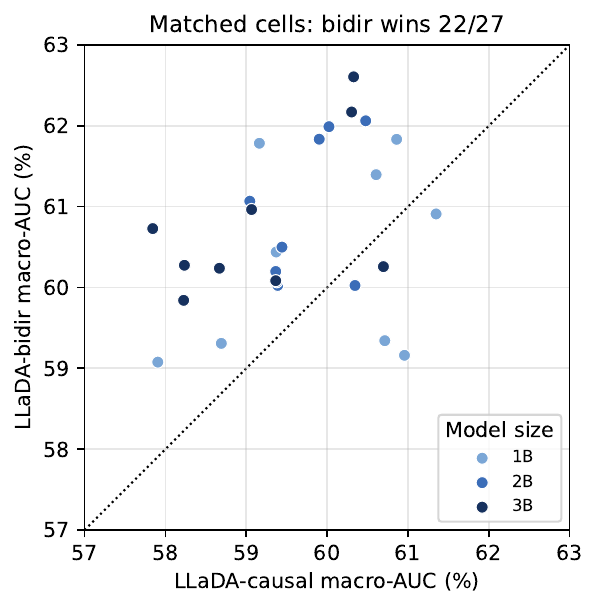}
\caption{Seed-matched macro-AUC comparison. Each point represents one pair of runs sharing the same model size, training subset, and seed. Points above the dotted diagonal favor the bidirectional mask.}
\label{fig:grid-scatter}
\end{figure}

\paragraph{The advantage is distributed across settings and seeds.}
Figure~\ref{fig:grid-scatter} plots the macro-AUC of every bidirectional run against that of its seed-matched causal run. Most points lie above the diagonal ($22$ of $27$ seed-matched run pairs), and the wins occur across all three model sizes, so the descriptive advantage is not confined to a single model scale.

\begin{figure}[t]
\centering
\includegraphics[width=0.92\columnwidth]{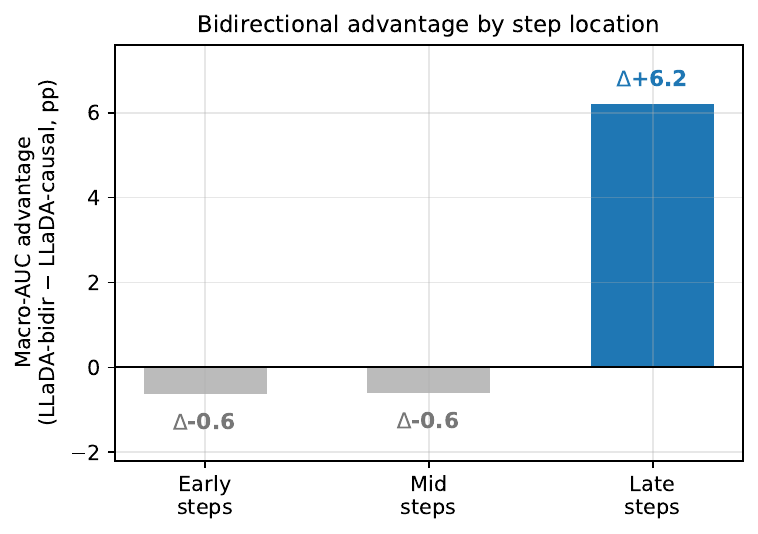}
\caption{Descriptive macro-AUC difference (LLaDA-bidir $-$ LLaDA-causal) by step location, pooled over the $27$ seed-matched run pairs. The early bucket contains no negative labels and therefore averages two present classes, whereas the middle and late buckets average three; values across buckets are not directly comparable.}
\label{fig:grid-positional}
\end{figure}

\paragraph{The positional decomposition shows its largest descriptive difference in the late bucket.}
Figure~\ref{fig:grid-positional} decomposes macro-AUC by step location. The bidirectional-minus-causal difference is $+6.2$ pp in the late bucket, whose absolute macro-AUC is low for both arms (bidir $23.8$, causal $17.6$). This comparison is descriptive: the early bucket contains no negative labels, so its macro-AUC averages two present classes, whereas the middle and late buckets average three. Consequently, cross-bucket differences are not directly comparable, and this decomposition does not isolate later-step evidence as the cause of the gain.

\begin{figure*}[t]
\centering
\includegraphics[width=0.48\textwidth]{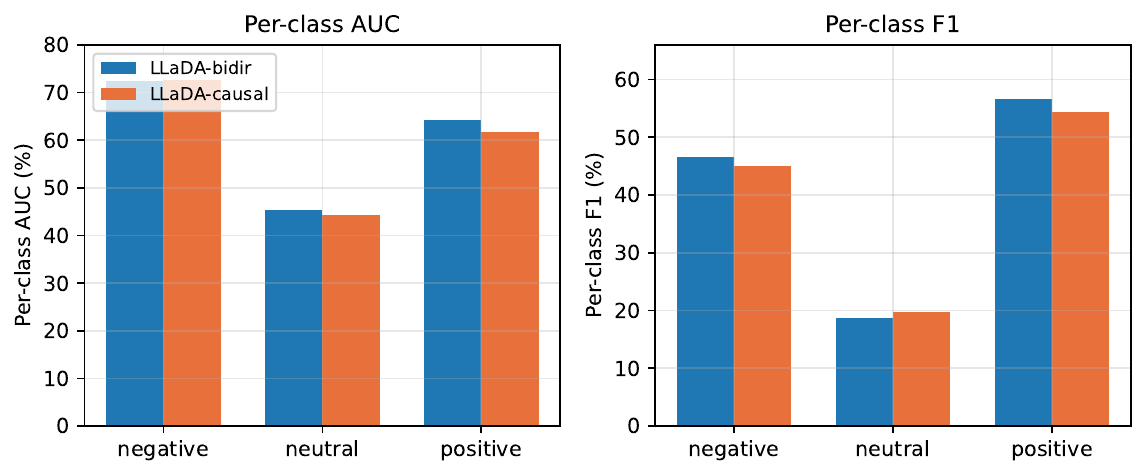}\hfill
\includegraphics[width=0.48\textwidth]{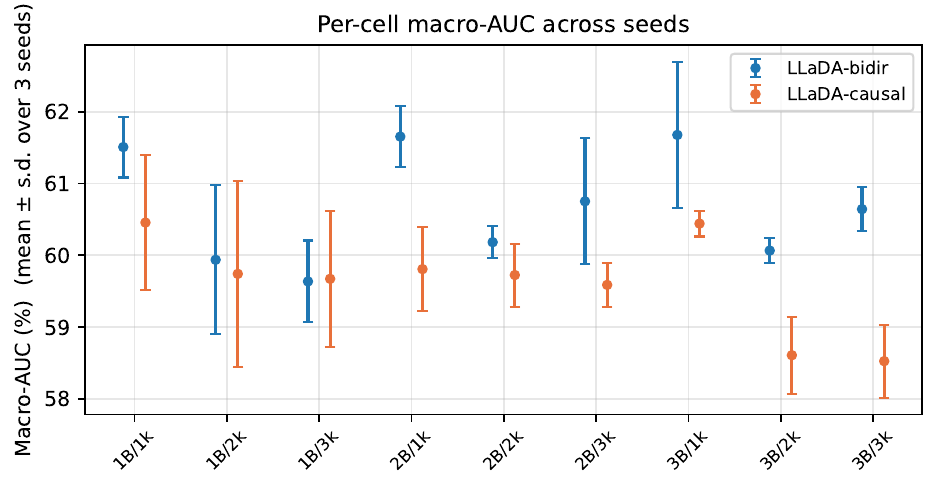}
\caption{Controlled-grid diagnostics. \textbf{Left:} per-class AUC and F1 for the negative, neutral, and positive step classes, averaged over the grid; both arms remain weak on the rare neutral class. \textbf{Right:} per-cell macro-AUC, mean $\pm$ standard deviation over the three seeds, for each (model size, training-set size) design cell.}
\label{fig:grid-diagnostics}
\end{figure*}

\paragraph{The aggregate difference is carried by the positive and negative classes.}
The left panel of Figure~\ref{fig:grid-diagnostics} breaks the metrics down by class. The bidirectional arm improves positive-class AUC and the negative- and positive-class F1, while negative-class AUC is essentially tied. The neutral class remains weak for both arms, with per-class AUC near $45$ and F1 below $20$. The classwise averages therefore do not indicate that the bidirectional mask resolves the neutral-class limitation.

\paragraph{Seed variation does not overturn the aggregate ranking.}
The right panel of Figure~\ref{fig:grid-diagnostics} shows the per-cell macro-AUC mean and spread across the three seeds. Although the seed bands overlap in some cells, the bidirectional mean is at or above the causal mean in eight of the nine (model size, training-set size) cells, and the separation is clearest at the largest model size.

\section{Label Imbalance and the Neutral Class}\label{sec:appendix_imbalance}
This appendix quantifies the PRM800K step-label imbalance discussed in the Limitations and reports the corresponding neutral-class diagnostics.

\begin{figure*}[t]
\centering
\includegraphics[width=0.73\textwidth]{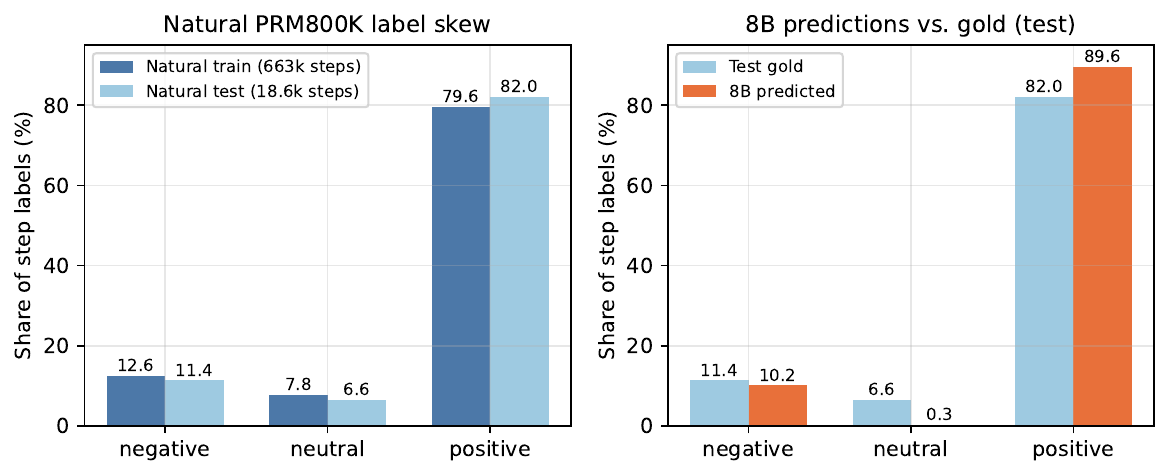}

\includegraphics[width=0.48\textwidth]{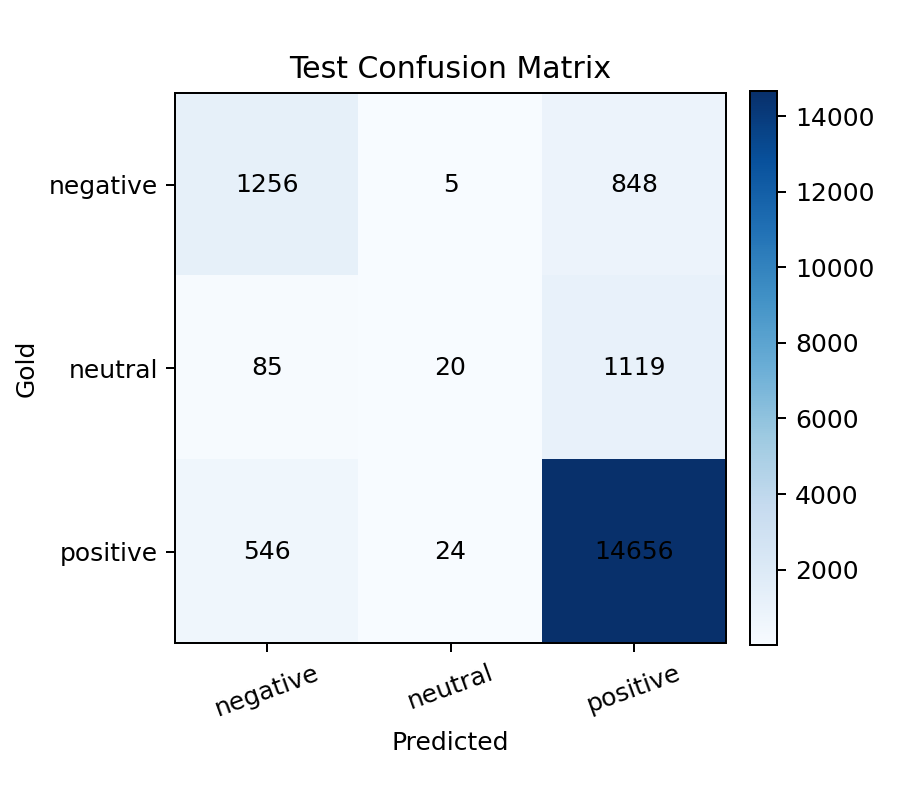}\hfill
\includegraphics[width=0.48\textwidth]{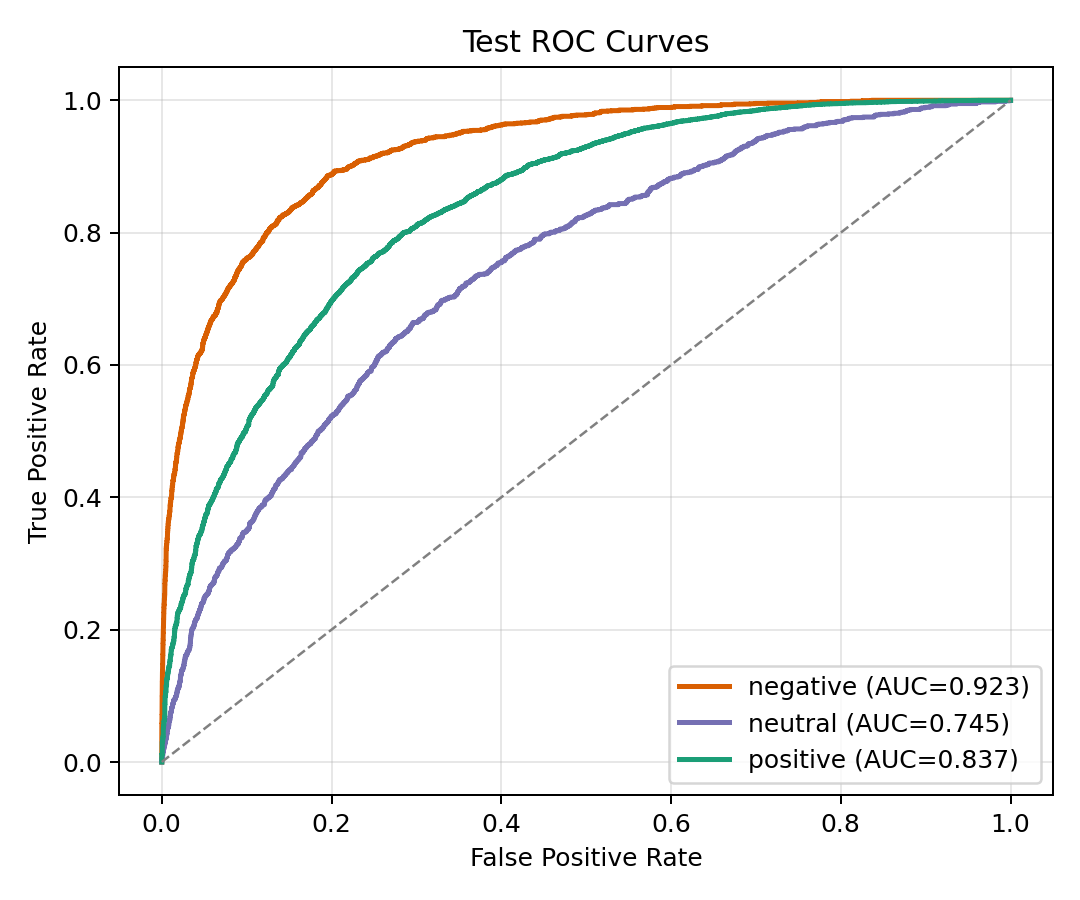}
\caption{PRM800K neutral-class diagnostics. \textbf{Top:} natural train/test label shares (left) and gold versus predicted test shares (right). \textbf{Bottom left:} the 8B model's confusion matrix; only $49$ of $18{,}559$ steps are predicted neutral, and most gold-neutral steps are routed to positive. \textbf{Bottom right:} one-vs-rest ROC curves; neutral-class AUC remains $0.745$ despite the collapsed argmax decisions.}
\label{fig:neutral-diagnostics}
\end{figure*}

\paragraph{Natural label distribution.}
The full PRM800K Phase~2 training set used for the 8B fine-tune contains $96{,}280$ traces with $663{,}101$ labeled steps; the label shares round to $79.6\%$ \textit{pos}, $12.6\%$ \textit{neg}, and $7.8\%$ \textit{neu}. The test split ($2{,}726$ traces, $18{,}559$ labeled steps) is similarly skewed: $82.0\%$ \textit{pos}, $11.4\%$ \textit{neg}, and $6.6\%$ \textit{neu}. Neutral is thus the rarest class.

\paragraph{Decision-rule sensitivity on the 8B model.}
The reported three-way per-class F1 uses the default argmax decision rule. Under this rule, \prm{} predicts \textit{neu} for only $49$ of $18{,}559$ PRM800K test steps ($0.3\%$, against a $6.6\%$ gold rate), yielding a neutral-class F1 of $3.1$. Because argmax has no explicit neutral threshold, this value characterizes the default three-way decision rule. To examine operating-point sensitivity, we additionally treat neutral detection as a one-vs-rest decision that predicts \textit{neu} whenever $p_i^{\mathrm{neu}} \geq \tau$. Under this rule, $\tau=0.5$ yields an F1 of $2.8$ (precision $41.9$, recall $1.5$), whereas applying the benchmark redundancy cutoff $\tau=0.15$ from Section~\ref{sec:method} yields an F1 of $25.7$ (precision $22.1$, recall $30.8$). The neutral-class AUC is $74.5$, confirming that useful ranking signal remains despite the default argmax decisions. Thus, the low argmax F1 should be interpreted as a decision-rule-specific result rather than as an absence of neutral-class signal; applications that require greater neutral recall call for a lower neutral-specific threshold and accept the corresponding precision trade-off. We report $\tau=0.15$ only as a sensitivity analysis; a deployment threshold must be selected on a separate validation set rather than on this test split.

\paragraph{Visualizing the imbalance and the collapse.}
Figure~\ref{fig:neutral-diagnostics} makes the imbalance and its consequence visible. The top panels show the natural label shares in train and test --- both about $80\%$ positive and under $8\%$ neutral --- and contrast the gold test shares with the model's predictions: the predicted neutral share collapses from a $6.6\%$ gold rate to $0.3\%$. The bottom panels show the same effect at the example level. The confusion matrix routes almost every gold-neutral step to the positive class, yet the one-vs-rest ROC curve still assigns the neutral class an AUC of $0.745$. Together, these views show the surviving neutral ranking signal (the ROC curve) alongside the collapsed argmax decisions (the confusion matrix).

\paragraph{Effect of balancing in the controlled grid.}
The 1B--3B grid instead trains on the balance-filtered pool described in Appendix~\ref{sec:appendix_repro}. Under the balance-filtered controlled-grid setup, neutral step F1 ranges from $14.8$ to $22.3$ across all 54 LLaDA grid runs, compared with the pretrained 8B model's $3.1$. Because these settings also differ in model scale, pretraining, data volume, and optimization, this comparison does not isolate the effect of balancing.

\section{Full MR-GSM8K Results (All Question Types)}\label{sec:appendix_gsm8k_full}
\begin{table*}[!t]
\centering\small
\setlength{\tabcolsep}{6pt}
\renewcommand{\arraystretch}{1.08}
\begin{tabular}{@{}lccccc@{}}
\toprule
\textbf{Model} & \textbf{Params} & \textbf{Sol-F1} & \textbf{Sol-AUC} & \textbf{Step-F1} & \textbf{Step-AUC} \\
\midrule
ReasonEval-Mistral & 7B & 61.2 & 75.6 & 61.9 & 84.9 \\
ReasonEval-WizardMath-V1.1 & 7B & 74.6 & 87.5 & 71.1 & 89.8 \\
WizardMath-V1.1 (control, eff\_bs=8) & 7B & 69.5 & 79.3 & 66.2 & 82.2 \\
ReasonEval-Llemma & 34B & 77.7 & 85.5 & 70.8 & 85.3 \\
Batch-matched LLaDA-bidir (eff\_bs=64) & 8B & 77.1 & 86.5 & 70.3 & 85.9 \\
LLaDA-PRM (eff\_bs=8, offline) & 8B & \textbf{87.9} & \textbf{94.0} & 78.7 & \textbf{91.6} \\
LLaDA-PRM (eff\_bs=8, online) & 8B & 86.7 & 93.3 & \textbf{78.9} & 90.9 \\
\bottomrule
\end{tabular}
\caption{Full MR-GSM8K results on all $2{,}999$ examples.}
\label{tab:gsm8k-full}
\end{table*}

The main MR-GSM8K comparison (Table~\ref{tab:benchmarks}) is reported on the original-question subset ($1{,}418$ examples), matching the split on which the published \citet{xia2024reasoneval} baselines are reported. Table~\ref{tab:gsm8k-full} repeats the comparison on the full MR-GSM8K benchmark ($2{,}999$ examples, spanning the \emph{original}, \emph{reversed}, and program-of-thought (\emph{POT}) question types). Baseline numbers are computed from ReasonEval's released per-example scores using our evaluation code; the same code reproduces their published original-subset numbers exactly (Table~\ref{tab:benchmarks}), so the full-set baseline values are directly comparable. For the headline \prm{}, all four metrics are lower on the full set than on the original-question subset, but the model remains above every published baseline on all four metrics. This confirms that its lead is not limited to the original-question subset.

\FloatBarrier

\section{Data and Reproducibility}\label{sec:appendix_repro}
This appendix documents the filtered dataset construction and statistical procedures used in the controlled experiments.

\subsection{Filtered training pool}
PRM800K's natural step-label distribution is heavily skewed toward \textit{pos}, and many traces never see \textit{neg} or \textit{neu} at all. For each trace, let $n_c$ be the number of labeled steps in class $c$ and define the balance ratio $\rho=\min_{c:n_c>0} n_c / \max_{c:n_c>0} n_c$, with $\rho=0$ when fewer than two classes are present. For the 1B--3B training pool, we retain Phase~2 traces with $\rho\geq0.5$ that contain at least one example of each of \textit{neg}, \textit{neu}, and \textit{pos}.

\newpage
\noindent This yields a $4{,}252$-trace filtered pool. We then randomly subsample $1{,}000$-, $2{,}000$-, and $3{,}000$-trace training subsets. We construct the evaluation set separately from the official PRM800K test split, applying only the $\rho\geq0.5$ criterion and obtaining the archived $608$-trace evaluation set.

\subsection{Statistical methodology}
Seed-level differences are computed within the 27 matched (model size, training-set size, seed) run pairs; win counts and mean differences are reported descriptively.
\end{document}